\documentclass[review,5p,twocolumn]{elsarticle}

\usepackage{amssymb}
\usepackage{amsmath}
\usepackage{subfigure}
\usepackage{mhchem}
\usepackage{float}
\usepackage{booktabs}

\journal{Nuclear Physics B}

\begin{document}

\begin{frontmatter}



\title{Toward Multi-Fidelity Machine Learning Force Field for Cathode Materials}


\author{Guangyi Dong, Zhihui Wang} 

\affiliation{organization={College of Computer Science and Artificial Intelligence, Fudan University},
            addressline={No. 2005, Songhu Road, Yangpu District},
            city={Shanghai},
            postcode={200438},
            state={Shanghai Municipality},
            country={China}}

\begin{abstract}
Machine learning force fields (MLFFs), which employ neural networks to map atomic structures to system energies, effectively combine the high accuracy of first-principles calculation with the computational efficiency of empirical force fields. They are widely used in computational materials simulations. However, the development and application of MLFFs for lithium-ion battery cathode materials remain relatively limited. This is primarily due to the complex electronic structure characteristics of cathode materials and the resulting scarcity of high-quality computational datasets available for force field training. In this work, we develop a multi-fidelity machine learning force field framework to enhance the data efficiency of computational results, which can simultaneously utilize both low-fidelity non-magnetic and high-fidelity magnetic computational datasets of cathode materials for training. Tests conducted on the lithium manganese iron phosphate (LMFP) cathode material system demonstrate the effectiveness of this multi-fidelity approach. This work helps to achieve high-accuracy MLFF training for cathode materials at a lower training dataset cost, and offers new perspectives for applying MLFFs to computational simulations of cathode materials.
\end{abstract}

\begin{keyword}


machine learning force field;  multi-fidelity; cathode material; graph neural network
\end{keyword}

\end{frontmatter}



\section{Introduction}
\label{sec1}

Lithium-ion battery technology has advanced rapidly in recent years, finding broad applications in consumer electronics, electric vehicles, and energy storage systems. The development of high-performance battery materials is crucial for achieving higher energy density and longer cycle life. In this context, computational modeling and simulation free from experimental constraints play an essential role in predicting and screening material properties\cite{1, 2, 3}.Traditional computational methods face a fundamental trade-off problem between accuracy and efficiency. While first-principles approaches such as density functional theory (DFT)\cite{4, 5} offer high accuracy and transferability, they are computationally expensive. In contrast, empirical force fields\cite{6,7} are efficient but often lack accuracy and generalizability across diverse chemical systems.Machine learning interatomic potentials (MLIPs) have emerged as a promising solution, combining the accuracy of quantum mechanical methods with the efficiency of classical force fields\cite{8,9,10}. 

In the early development of MLIPs, models often used carefully designed descriptors—such as atom-centered symmetry functions (ACSF)\cite{11}, smooth overlap of atomic positions (SOAP)\cite{12}, and atomic cluster expansion (ACE)\cite{13}—to capture symmetries and characterize the local chemical environment of atoms. These were combined with fitting networks to predict energy, forming the so-called local descriptor-based MLIPs (L-MLIPs). Typical examples include moment tensor potentials (MTP)\cite{14}, GAP\cite{15}, DeePMD\cite{16,17,18}, and the ANI series\cite{19,20,21}, among others. These models offer high representational efficiency and can finely describe local chemical environments, but their universality across chemical systems (especially across different element types) remains limited.Recently, leveraging the inherent similarity between the atomic structures of materials and graph structures\cite{22}, methods that use graph neural networks to learn interatomic interaction potential energy surfaces have emerged, known as graph neural network-based MLIPs (G-MLIPs). They exhibit strong universality and have become the mainstream architectural choice for large-scale general-purpose pre-trained force fields, such as M3GNet\cite{23}, DimeNet\cite{24,25}, Sevennet\cite{26}, and MACE\cite{27}, among others.

A major bottleneck in training high-fidelity MLIPs is the scarcity of high-quality reference data\cite{28}. High-accuracy computational methods (e.g., DFT with hybrid functionals or coupled cluster methods) are prohibitively expensive, limiting the size of training datasets\cite{29,30}. 

Due to the high cost of high-quality training datasets, researchers aim to improve data efficiency for training with small datasets, mainly through sampling optimization (e.g., active learning frameworks like DP-GEN\cite{31}, or new strategies to cover configuration space\cite{32}) and training method innovations (e.g., $\Delta$-learning-based DPRc\cite{33}, transfer learning-based MEGNet\cite{34}, and meta-learning\cite{35}). However, these methods have their own limitations: active learning involves many iterations increasing computational cost; $\Delta$-learning requires paired low- and high-fidelity data; transfer learning faces catastrophic forgetting or negative transfer problems. 

Besides, multi-fidelity machine learning methods at the model level have emerged\cite{36}, synergizing data of different fidelities (low-fidelity data for broad configuration space and a small amount of high-fidelity data for calibrating critical regions) to build high-accuracy models, with applications in G-MLIPs like SevenNet-MF\cite{30}, MEGNet\cite{37} and M3GNet\cite{38} multi-fidelity frameworks.Currently, MLIPs see limited application in computational simulation of lithium battery cathode materials; most MLIP simulations of battery materials focus on electrolytes\cite{39} or anodes\cite{40}. There are several reasons for these challenges. Cathode materials possess complex electronic structures, primarily due to transition metal oxidation state changes that occur during lithiation and delithiation. These changes lead to intricate arrangements of magnetic moments, which make it difficult for machine learning interatomic potentials (MLIPs) to accurately represent the potential energy surfaces. Additionally, this electronic complexity causes density functional theory (DFT) calculations to be hard to converge, resulting in a scarcity of high-precision data. Furthermore, open datasets often use inconsistent methods and standards, which hinders their direct use. Thus, developing high-data-efficiency MLIP models via multi-fidelity machine learning for cathodes is key to promoting their practical use in lithium battery materials.

To address this, we introduce a multi-fidelity graph-based MLIP framework based on CHGNet\cite{41}, which incorporates atomic magnetic moments and is thus well-suited for cathode materials. Our approach integrates data from different levels of theory, using non-magnetic calculations to circumvent convergence issues while incorporating magnetic reference data where available. We demonstrate the effectiveness of this approach on a lithium manganese iron phosphate cathode material, highlighting its potential for efficient and accurate modeling of complex battery materials.

\section{Architecture of Multi-Fidelity Graph Machine Learning Force Field}
\label{sec:framework}

This framework is built upon the CHGNet model\cite{41}, a graph neural network model that expresses the interatomic potential energy function. By constructing a crystal graph composed of an atom graph (with atoms as nodes and bonds as edges) and a bond graph (with bonds as nodes and bond angles as edges) to represent the crystal structure, and iteratively updating the features of atoms, bonds, and bond angles through message passing, CHGNet efficiently expresses and fits complex interatomic interactions in materials. The CHGNet model primarily consists of modules such as graph transformers, embedding layers, message passing layers, and readout layers. The input to the model is the crystal structure, including lattice vectors, elemental symbols, and coordinates of atoms. The output includes the total energy of a unit cell ($E_\text{tot}$), the forces on each atom ($\mathbf{F}_i$), the lattice stress ($\mathbf{\sigma}$), and the magnetic moments of each atom ($m_i$). Similar to other MLIPs, CHGNet expresses the total energy of the system as the sum of atomic contributions, where each atomic contribution is a function of its local environment. The expression is as follows:
\begin{equation}
    E_\text{tot} = E_\text{c} + \sum_i L_{\text{readout}}(\mathbf{v}_i^{n_\text{L}}) ,
\end{equation}
where $E_\text{c}$ is the contribution that depends solely on the atomic composition of the material and is independent of the atomic structure; $\mathbf{v}_i^{n\text{L}}$ is the feature of the $i$-th atom after $n_\text{L}$ layers of message passing (or graph convolution) operations. $L_\text{readout}$ denotes the fitting network of the readout layer. After obtaining the total energy through forward propagation, the atomic forces and lattice stress are derived by automatic differentiation of the total energy with respect to the atomic coordinates ($\mathbf{r}_i$) and lattice strain ($\mathbf{\epsilon}$), respectively.
\begin{equation}
\mathbf{F}_{i} =-\frac{\partial E_{\text{tot}}}{\partial \mathbf{r}_{i}},\quad
\mathbf{\sigma}  = \frac{1}{V} \frac{\partial E_{\text{tot}}}{\partial \mathbf{\epsilon}} .    
\end{equation}
where $V$ is the volume of the material's unit cell, and stress can be viewed as the force acting on the degrees of freedom of the lattice vectors. In addition to these three common outputs of machine learning force fields, the CHGNet model derives the magnetic moment of each atom from the atomic features of the penultimate layer ($\mathbf{v}i^{n\text{L}-1}$) through a fitting network ($L_\text{m}$):
\begin{equation}
    m_{i} = L_{\text{m}}\left(\mathbf{v}_{i}^{n_\text{L}-1}\right) . 
    \label{eq:magmom_def}
\end{equation}
By predicting and constraining the magnetic moments of transition metal atoms, the CHGNet force field enables the differentiation of various oxidation states of metal atoms, making it particularly suitable for computational simulations of lithium-ion battery cathode materials.

Building upon the CHGNet model, this work introduces a multi-fidelity learning mechanism to construct a multi-fidelity graph machine learning force field architecture. This framework enables the simultaneous utilization of data generated from multiple accuracy levels or computational methods, thereby improving data efficiency. Fidelity is represented by an integer $f$ ranging from 1 to $n_\text{F}$, where $n_\text{F}$ denotes the number of distinct fidelity types. This integer characterizes the data quality or "fidelity" of the dataset. In practical implementation, different $f$ values can be assigned based on the data quality of the dataset. For instance, higher data quality corresponds to a larger $f$.

To support the multi-fidelity learning mechanism, this work modifies four components of the original CHGNet architecture:\\
1) Implementing fidelity-dependent atomic embeddings in the embedding layer.\\
2) Enabling fidelity-dependent message generation in the message passing layer.\\
3) Incorporating fidelity-dependent fitting networks in the readout layer.\\
4) Introducing a fidelity-dependent composition model outside the graph neural network.

\begin{figure*}[htbp]
\centering
\subfigure[]{
\includegraphics[width=0.45\textwidth]{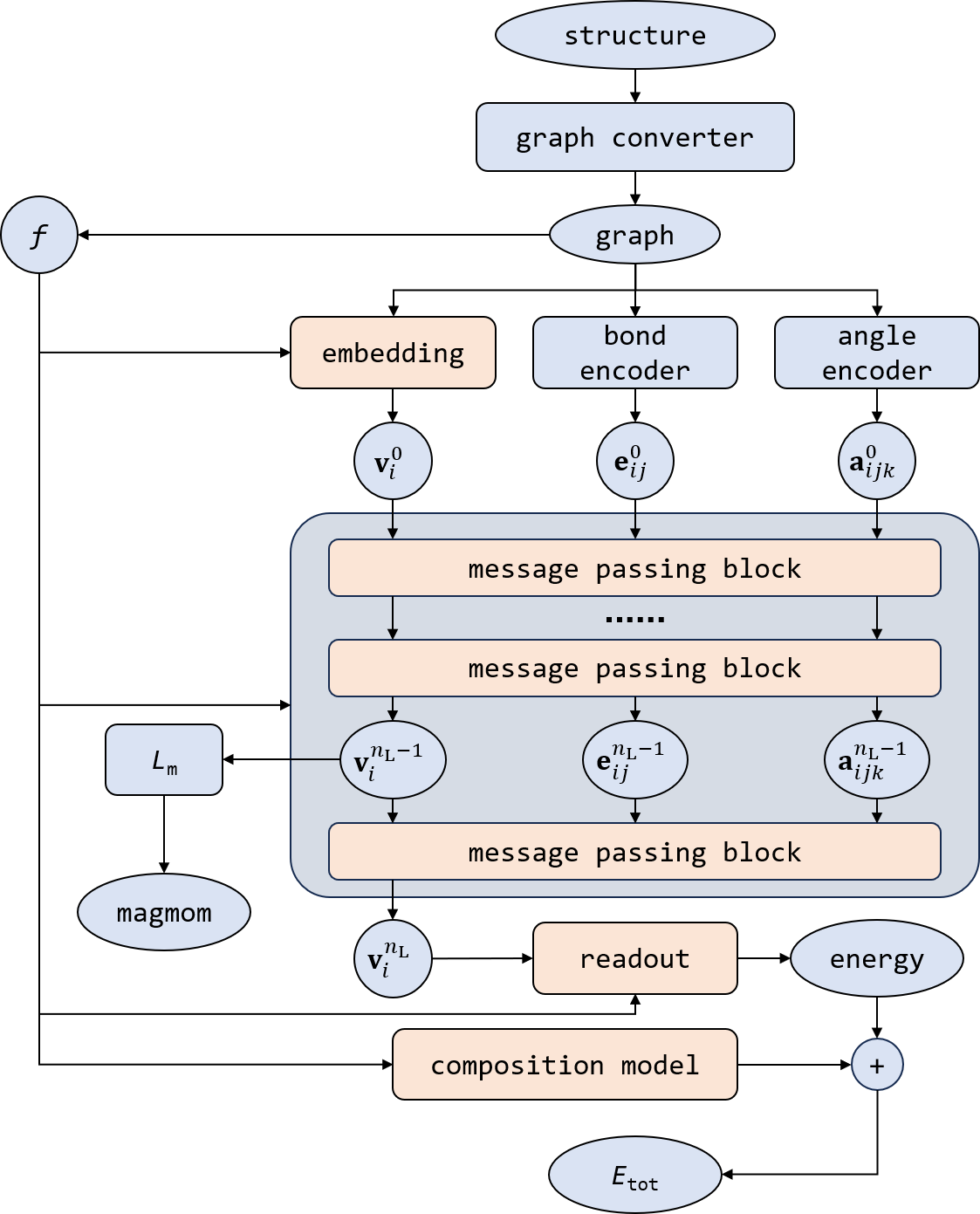}}
\label{fig:whole_structure}
\hfill
\subfigure[]{
\includegraphics[width=0.45\textwidth]{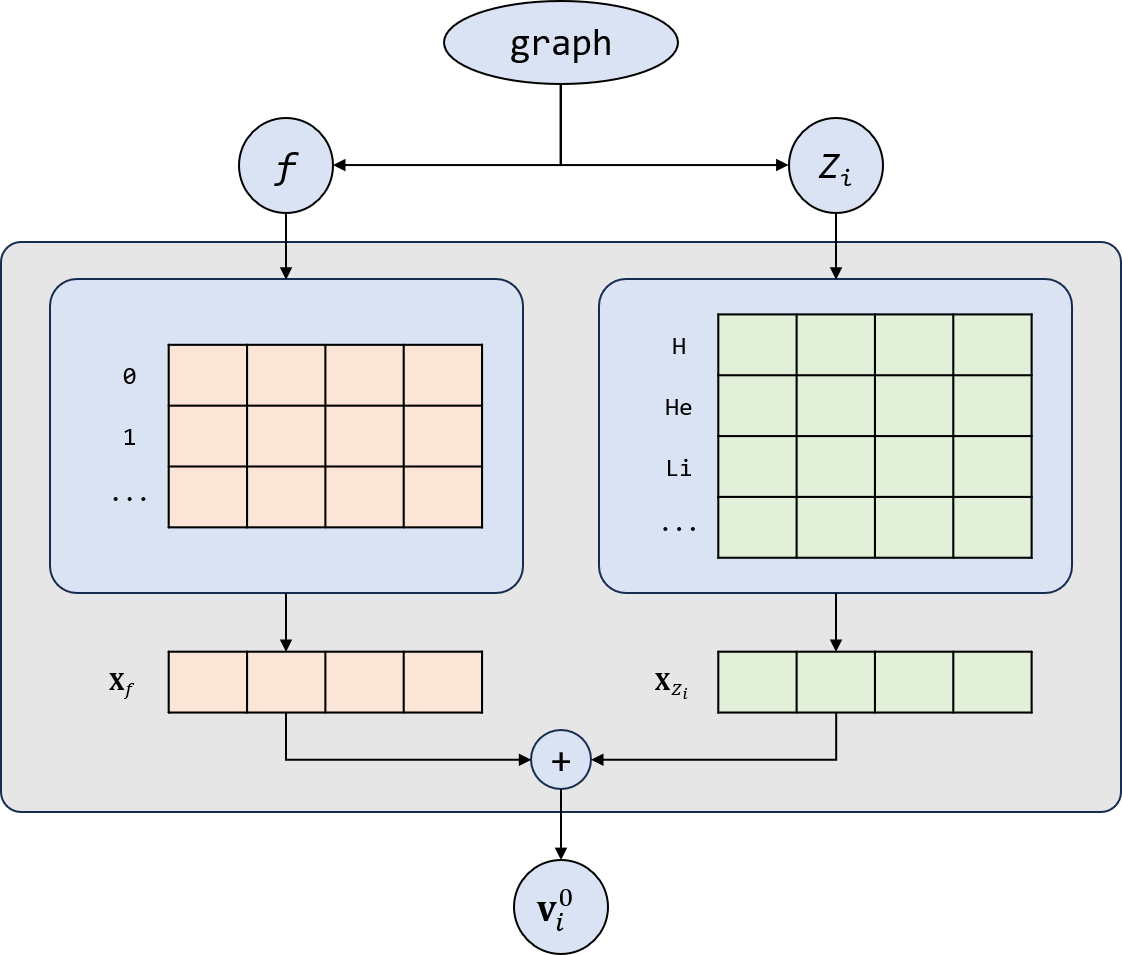}}
\label{fig:embedding}
\hfill
\subfigure[]{
\includegraphics[width=0.45\textwidth]{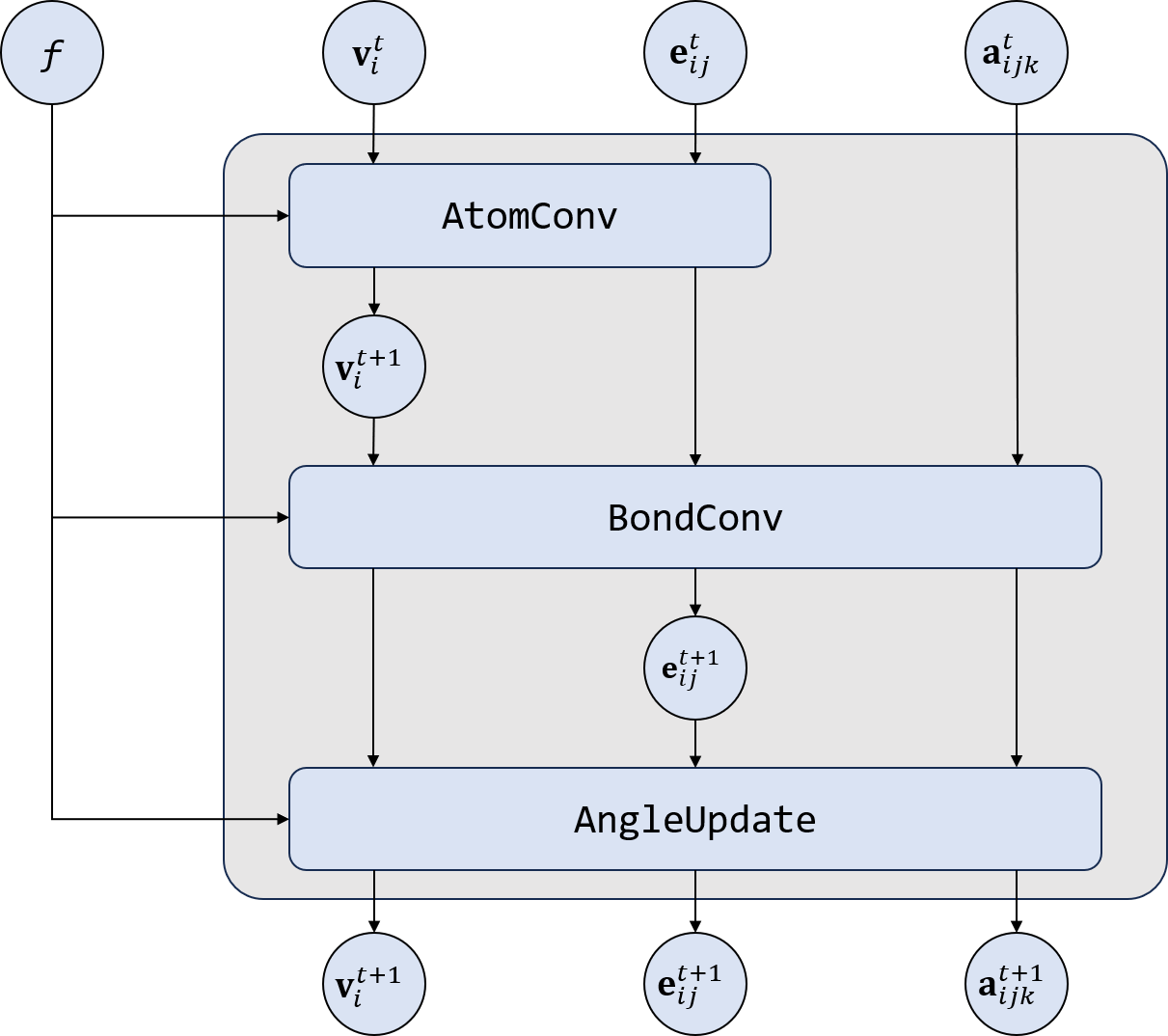}}
\label{fig:graph_conv}
\hfill
\subfigure[]{
\includegraphics[width=0.45\textwidth]{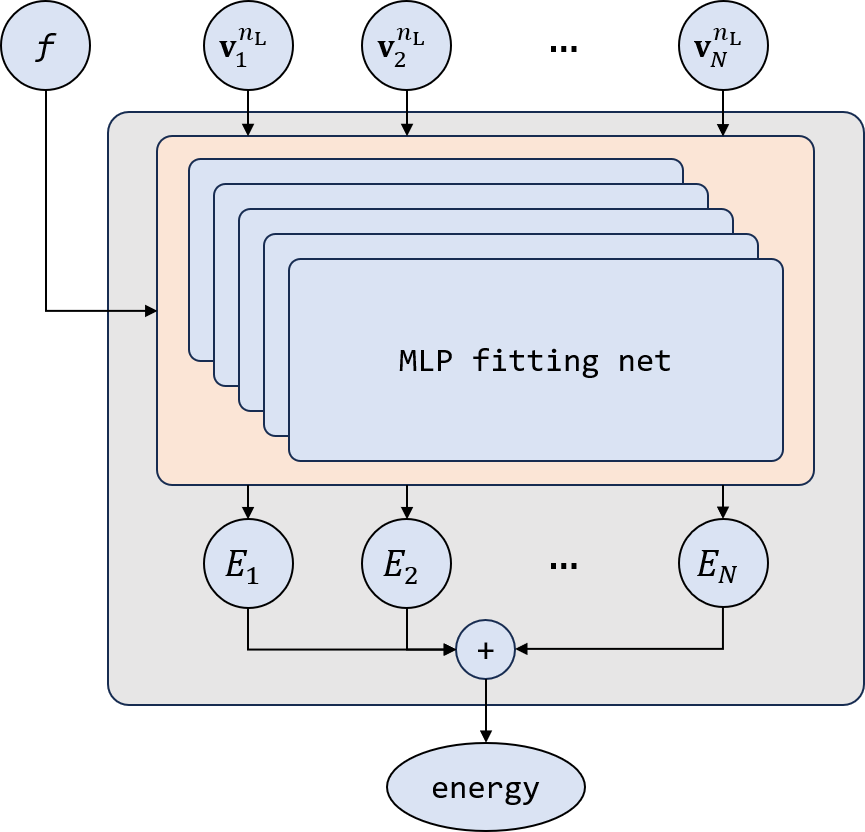}}
\label{fig:fitting_net}
\caption[MLFF]{
Schematic diagram of the multi-fidelity graph machine learning force field architecture. (a) Overall architecture; (b) fidelity-dependent atomic embedding layer; (c) fidelity-dependent message-passing layer; (d) fidelity-dependent readout layer. In (a), the modules related to fidelity are indicated in orange.
}
\label{multi-fidelity}
\end{figure*}

The overall structure of the multi-fidelity graph machine learning force field is illustrated in Figure \ref{multi-fidelity}(a).

Figure \ref{multi-fidelity}(b) illustrates the fidelity-dependent atomic embedding. The embedding layer converts the atomic number $Z_i$ into a fixed-dimensional embedded feature vector, which serves as the initial feature $\mathbf{v}_i^0$ of node $i$. To achieve differentiation between data of different accuracy levels, the embedding vector is designed to depend on both $Z_i$ and $f$. The expression is as follows:
\begin{equation}
    \mathbf{v}^0_i = \mathbf{X}_{Z_i} + \mathbf{X}_f .
\end{equation}
That is, we express the embedding vector as the sum of the atomic number embedding $\mathbf{X}_{Z_i}$ and the fidelity embedding $\mathbf{X}_f$. The advantages of this approach are as follows: On the one hand, we introduce the distinction of data fidelity at the atomic embedding level, which is equivalent to treating the same element in datasets of different precisions as "atoms with different properties"—thus having different interactions—and which can therefore be used to represent potential energy surfaces of different precisions. On the other hand, introducing the fidelity embedding vector $\mathbf{X}_f$ in an additive form ensures that, for any two different elements, the difference between their embedding vectors remains the same across data of different precisions. This thereby increases the possibility of mutual transfer and learning between data of different fidelities.

Figure \ref{multi-fidelity}(c) shows the fidelity-dependent message-passing layer, or graph convolutional layer. First, one-hot encoding is performed on the fidelity $f$: $\mathbf{f}_g= \text{one\_hot}(f,n_\text{F})$, which is a column vector of length $n_\text{F}$ where all entries are 0 except for the $f$-th row, which is 1:
\begin{equation}
    \mathbf{f}_g = \left(0,\cdots, 1, \cdots, 0\right)^\text{T}, \quad (\mathbf{f}_g)_i = \delta_{if} .
\end{equation}
where g denotes the dependence on the crystal graph. Inserting $\mathbf{f}_g$ into the neural network can introduce additional parameters to achieve dependence on fidelity. For example, for the basic linear layer unit
\begin{equation}
    \mathbf{a}^{(k)} = \mathbf{W}^{(k,k-1)}\mathbf{a}^{(k-1)} + \mathbf{b}^{(k)},
\end{equation}
After concatenating \(\mathbf{f}_g\) with the input \(\mathbf{a}^{(k-1)}\), the function of this unit becomes:
\begin{equation}
    \mathbf{a}^{(k)} = \left[ \mathbf{W}^{(k,k-1)}, \mathbf{W}^{(k,k-1)}_{\text{F}} \right] \left[
    \begin{matrix}
        \mathbf{a}^{(k-1)} \\
        \mathbf{f}_g
    \end{matrix}
    \right] + \mathbf{b}^{(k)}.
\end{equation}
Here, $\mathbf{W}_\text{F}^{(k,k-1)}$ is a newly added parameter matrix, where each column corresponds to a fidelity and is only activated when training with data of the corresponding precision.

In this work, $\mathbf{f}_g$ is inserted at the entrances of the three submodules in the CHGNet graph convolutional layer: AtomConv (atomic graph convolution), BondConv (bond graph convolution), and AngleUpdate (angle update). This enables them to generate fidelity-dependent messages for transmission, thereby updating the features of atoms, bonds, and bond angles.
\begin{equation}
    \begin{aligned}
        \mathbf{m}_{v_i}^{t}&=\phi_{\mathrm{v}}^{t}(\mathbf{v}_{i}^{t} \oplus \mathbf{v}_{j}^{t} \oplus \mathbf{e}_{i j}^{t} \oplus \mathbf {f}_{g}),\\
        \mathbf{m}_{e_{jk}}^{t}&={\phi}_{\mathrm{e}}^{t}(\mathbf{e}_{i j}^{t}\oplus \mathbf{e}_{j k}^{t}\oplus \mathbf{a}_{i j k}^{t} \oplus \mathbf{v}_{j}^{t+1}\oplus \mathbf {f}_{g}),\\
        \mathbf{m}_{a_{ijk}}^{t}&={\phi}_{\mathrm{a}}^{t}(\mathbf{e}_{i j}^{t+1}\oplus\mathbf{e}_{j k}^{t+1} \oplus\mathbf{a}_{i j k}^{t} \oplus \mathbf{v}_{j}^{t+1}\oplus \mathbf {f}_{g}).
    \end{aligned}
\end{equation}
Here, $\phi$ is a gated multi-layer perceptron (gated MLP). $\mathbf{v}_{i}^{t}$, $\mathbf{e}_{i j}^{t}$, and $\mathbf{a}_{i j k}^{t}$ are the feature vectors corresponding to atom $v_i$ at step $t$, the bond $e_{ij}$ formed by atoms $v_i$ and $v_j$, and the bond angle $a_{ijk}$ formed by bonds $e_{ji}$ and $e_{jk}$, respectively. $\oplus$ denotes vector concatenation. 

The architectural adjustment in this part introduces fidelity into the message generation stage. Through multiple graph convolution operations, it can describe the fine differences in the potential energy surfaces of interatomic interactions corresponding to different computational methods or data of different fidelities via a small number of additional parameters, while retaining most of the shared parameters of the original architecture to describe the common overall trends of different potential energy surfaces as structures change.

Figure \ref{multi-fidelity}(d) shows the fidelity-dependent fitting network. After going through $n_\text{L}$ message passing layers, the final feature vector $\mathbf{v}_i^{n_\text{L}}$ for each atom is obtained. Subsequently, through different fitting networks according to different fidelities $f$, the total energy of the unit cell is obtained:
\begin{equation}
     E_\text{tot} = E_\text{c}^f + \sum_i L_{\text{readout}}^f(\mathbf{v}_i^{n_\text{L}}) .
     \label{eq:e_sum}
\end{equation}
Here, $L^f_{\text{readout}}$ is the readout layer corresponding to fidelity $f$, and it is one of the $n_\text{F}$ copies of the MLP readout layer in the original CHGNet architecture. After multiple graph convolutions or message passings in the Graph Neural Network, the resulting atomic feature vectors serve as excellent descriptors of their local environments. Here, we assume that the structural descriptions required for the potential energy surfaces corresponding to different computational methods may be relatively similar, while the mapping from descriptors to energy may vary significantly. Therefore, using different fitting networks according to fidelity $f$ can effectively describe the major differences in potential energy surfaces.

Additionally, as shown in Equation (\ref{eq:e_sum}), we use a fidelity-dependent composition model $E_\text{c}^f$. In the main part of the machine learning force field, we employ a graph neural network to describe the complex structure-dependent interatomic interaction potential energy surfaces. However, the total energy values generated by first-principles calculations usually depend on the energy zero point of each atom. This part of the energy typically contributes significantly to the absolute value of the total energy, and the reference energies of atoms for each element vary with different computational methods. Thus, datasets generated by different computational methods are not suitable for direct joint training. The composition model is used to describe this part of the energy contribution, which is independent of the relative positions of atoms; it is a linear function of the elemental composition of the material:
\begin{equation}
    E_\text{c}^f =\sum _{a=1}^{94} w_{a,f}\cdot  x_a \quad .
    \label{eq:composition_model}
\end{equation}
Here, \(x_a\) is the atomic fraction of element a in the material, and \(w_{a,f}\) is the weight for element a corresponding to fidelity $f$. These weight parameters are obtained through linear fitting of the total energy of the unit cells in the dataset with fidelity $f$ using Equation (\ref{eq:composition_model}). Data with different fidelities correspond to different composition models, enabling the graph neural network, which constitutes the main part of the machine learning force field to focus on describing interactions. This thereby enhances the expressive and predictive capabilities of the overall model.

\section{Multi-fidelity Force Field Training Experiments}

In this section, force field training tests are conducted on practical lithium battery cathode material systems to verify the effectiveness of the multi-fidelity architecture in improving data utilization efficiency. First, by varying the composition of the training set, the effects of training with a mixture of low-quality and high-quality datasets are compared with those of training with only a single high-quality dataset, aiming to demonstrate the effectiveness of the multi-fidelity architecture proposed in this work. Second, through ablation experiments, the impact of each fidelity-dependent module on model performance is examined to confirm the necessity of each module. Finally, this section also compares this architecture with common transfer learning strategies to demonstrate the superior performance of the multi-fidelity model.

The dataset used in the experiment is a dataset of lithium iron manganese phosphate (\ce{Li_xMn_yFe_{1-y}PO4}, $0 \leq x,y \leq 1$, LMFP) cathode materials generated from DFT-based first-principles calculations. A $1 \times 1 \times 2$ supercell was adopted, with random partial delithiation and random mixing of \ce{Mn} and \ce{Fe} atoms. In the calculations, the PBE functional\cite{29} was used to describe the exchange-correlation effects of electrons, and Hubbard \textit{U} corrections of 4.5 eV and 4.3 eV were added to the d-orbitals of \ce{Mn} and \ce{Fe} atoms, respectively\cite{42} to reduce the delocalization error of this functional in describing transition metal systems. Two types of calculations were performed based on whether magnetism was explicitly considered, generating datasets of two precisions. First, spin-polarized magnetic DFT calculations were conducted, explicitly considering the magnetic moments of \ce{Mn} and \ce{Fe} atoms, resulting in a high-quality dataset with a total of 1911 structure frames, with 10\% reserved for testing. Subsequently, spin-unpolarized non-magnetic DFT calculations were performed, obtaining a low-quality dataset with 1736 structure frames. Compared with the magnetic dataset, the non-magnetic dataset calculations used a reduced energy cutoff of the plane-wave basis set was reduced from 520 eV to 400 eV, and the spacing of $\mathbf{k}$-point sampling in the reciprocal space was increased from $0.2 \,\mathrm{\AA}^{-1}$ to $0.35\,\mathrm{\AA}^{-1}$, thus yielding lower calculation precision. A more significant difference is that the data in this non-magnetic dataset lacks atomic magnetic moment labels.

The optimized hyperparameters used in this experiment are listed in Table \ref{tab:hyperpars}. The task objectives EFSM (energy, force, stress, magnetic moment) represent the four targets respectively, and the task weights correspond to these four objectives respectively.To characterize the randomness in both the selection of structural data and the process of the force field model, all experiments were repeated three times under the same hyperparameter settings, with the test results averaged and uncertainties calculated.

\begin{table}[H]
    \centering
    \caption{Hyperparameters for training the multi-fidelity machine learning force field of the LMFP material system}
    \begin{tabular}{p{4cm}p{3cm}}
        \toprule
        Hyperparameter                 &     Set value \\
        \midrule
        Batch size              & 4             \\ 
        Epoch                 & 60            \\ 
        Learning rate                 & 0.005         \\ 
        Feature dimension               & 64            \\ 
        Number of convolution layers (\(n_\text{L}\))              & 4             \\ 
        Hidden layer dimension                & 64            \\ 
        Number of radial basis functions    & 31               \\
        Number of angular basis functions    & 31               \\
        Optimizer           & Adam             \\
        Learning rate scheduler      & CosLR            \\
        Loss function         & Huber            \\
        Task target         & efsm             \\
        Target weights         & 1:1:0.1:0.1      \\ 
        \bottomrule
    \end{tabular}
    \label{tab:hyperpars}
\end{table}

\subsection{Validation of Multi-Fidelity Architecture Effectiveness}
\label{sec:effectiveness}

\begin{figure*}[htb]
\centering
\includegraphics[width=\textwidth]{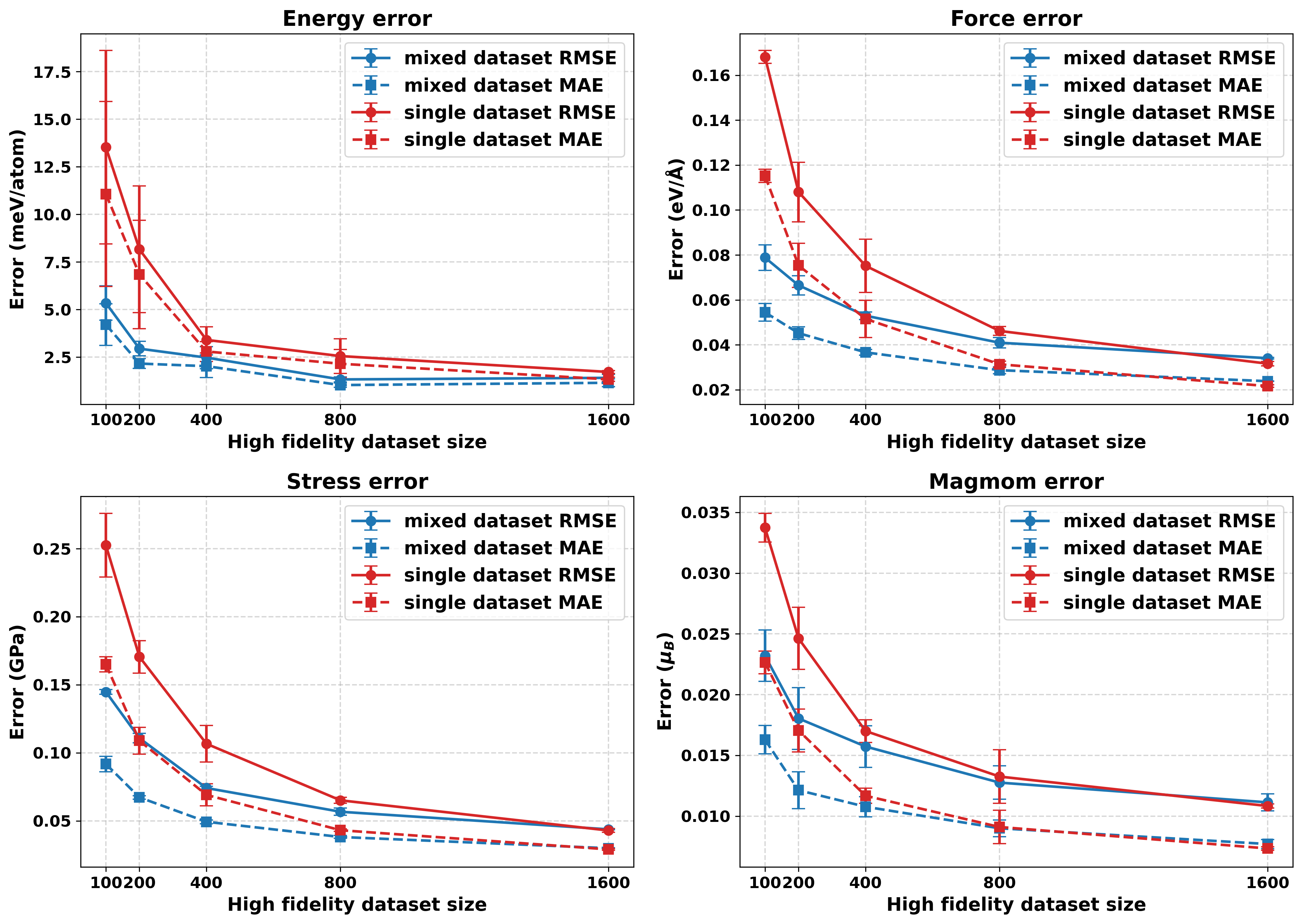}
\caption[experiment1-1]{
Comparison of test errors between force fields trained using a mixed dataset for multi-fidelity force field training and those trained using only a single high-fidelity dataset. The mixed dataset includes all low-fidelity data and a certain number of high-fidelity data. The abscissa in the figure represents the number of high-fidelity data.
}
\label{fig:MF_effects}
\end{figure*}

Figure \ref{fig:MF_effects} shows the improvement in prediction accuracy brought by force field training using the multi-fidelity architecture. The figure compares the magnitude of test errors between multi-fidelity training with a mixed dataset (including all low-fidelity data and a certain number of high-fidelity data) and training with a single high-fidelity dataset. First, as can be seen from Figure \ref{fig:MF_effects}, with the increase in high-fidelity training data, the errors of all four types of labels (total energy, atomic forces, lattice stress, and atomic magnetic moments) on the test set decrease significantly. This is consistent with the general expectations for force field training, partially verifying the correctness of the architecture implementation. Second, whether evaluated by mean absolute error (MAE) or root mean square error (RMSE), the learning curves of the mixed dataset generally lie below those of the single dataset. This demonstrates that with the proposed multi-fidelity training architecture in this work, low-fidelity datasets can aid training to improve the prediction accuracy of high-fidelity labels. This improvement in accuracy is particularly pronounced when high-fidelity data is relatively scarce (e.g., fewer than 400 structure frames). Additionally, notably, even though the low-fidelity data lacks magnetic moment labels, the mixed dataset still leads to an improvement in the prediction accuracy of magnetic moment labels. As indicated by Equation (\ref{eq:magmom_def}), this stems from the improved descriptive accuracy of the internal node features $\{\mathbf{v}_i^{n_\text{L}-1}\}$ of the graph neural network, further confirming that low-fidelity data, by providing information on the general behavior of interatomic interactions during training, ultimately enhances the predictive ability for high-fidelity labels.

\begin{figure*}[htbp]
\centering
\subfigure[]{
\includegraphics[width=0.8\textwidth]{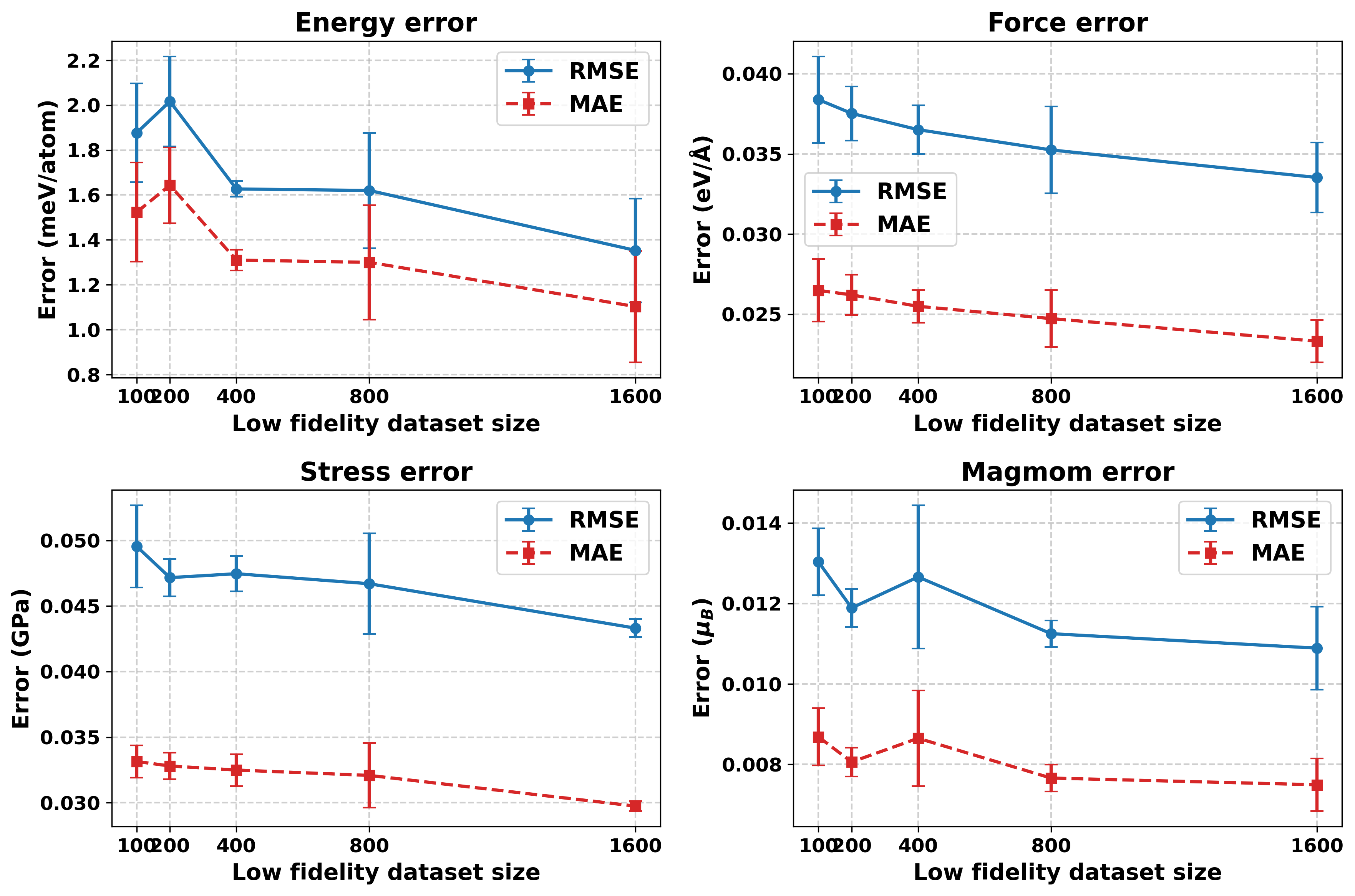}}
\label{fig:lf_size_hf400}
\centering
\subfigure[]{
\includegraphics[width=0.8\textwidth]{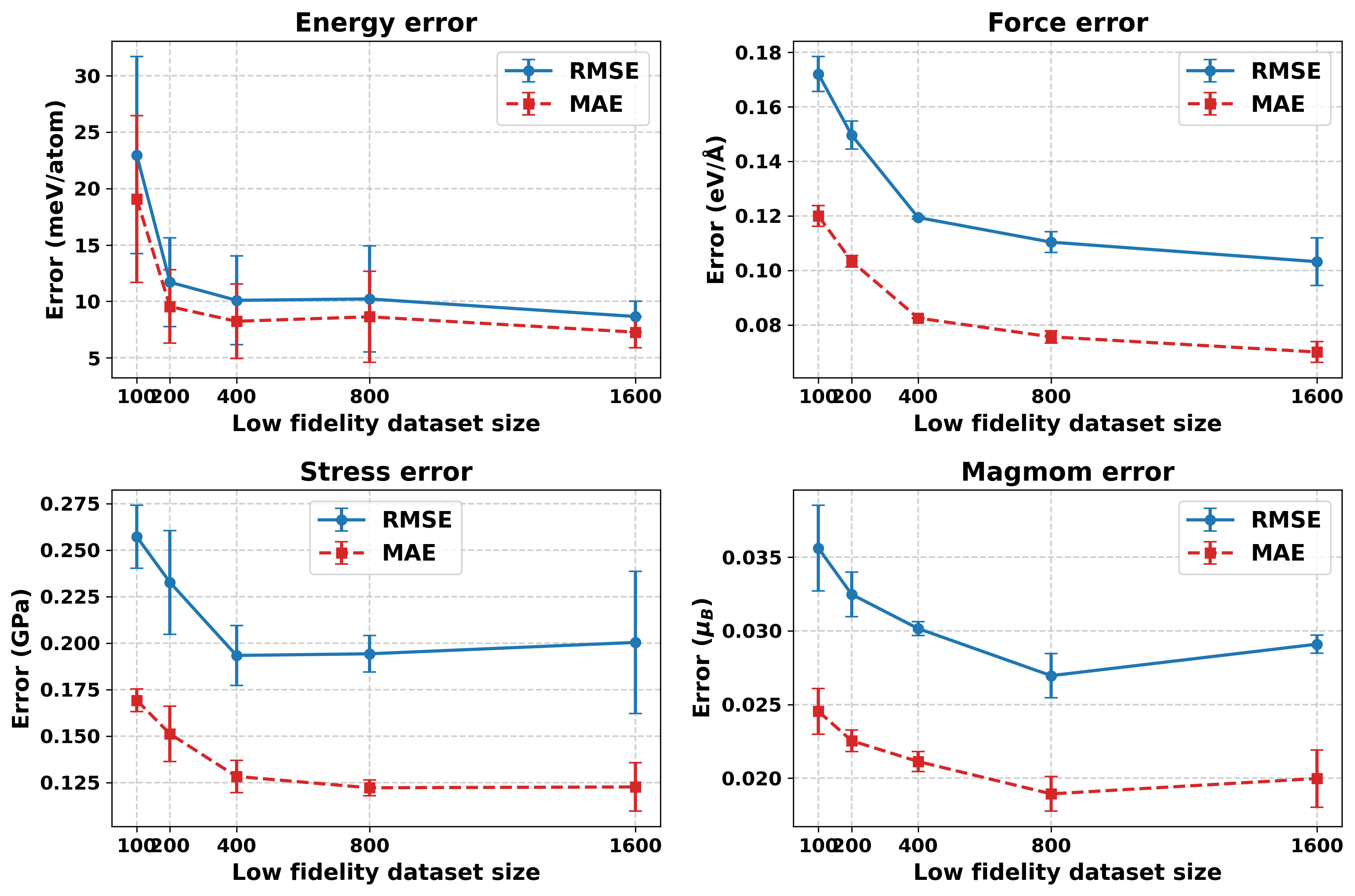}}
\label{fig:lf_size_hf50}
\caption[experiment1-2]{The impact of the size of the low-fidelity dataset in the mixed dataset for multi-fidelity training on the test error of high-fidelity labels. (a) The number of high-fidelity data is fixed at 400; (b) The number of high-fidelity data is fixed at 50.}
\label{fig:Low_fidelity_sizes}
\end{figure*}

Figure \ref{fig:Low_fidelity_sizes} shows the impact of the size of the low-fidelity dataset on the model's predictive performance when conducting multi-fidelity training using a mixed dataset. As can be seen from the figure, when the number of high-fidelity data is fixed at 400 (Figure \ref{fig:Low_fidelity_sizes}(a)) and 50 (Figure \ref{fig:Low_fidelity_sizes}(b)) respectively, the overall predictive accuracy of the model gradually increases with the increase in the number of low-fidelity data, especially for the atomic force labels, which are most critical for the force field. This indicates that the multi-fidelity force field model indeed learns information on the interatomic interactions of materials from low-fidelity data: the more data there is, the more information is learned.

\subsection{Functional Module Testing}

In this section, ablation experiments are used to test the impact of each fidelity-dependent functional module on the training effect of the multi-fidelity force field model, and to verify the functional necessity of different modules. For convenience, the four submodules proposed in Section \ref{sec:framework}—the fidelity-dependent atomic embedding layer (\textbf{E}mbedding), message passing block (\textbf{M}essage passing block), readout layer (\textbf{R}eadout), and composition model (\textbf{C}omposition model)—are denoted by the symbols \textbf{E}, \textbf{M}, \textbf{R}, and \textbf{C}, respectively. A single symbol indicates that only the corresponding fidelity dependence is introduced in the multi-fidelity architecture; the symbols $\bar{\text{\textbf{E}}}$, $\bar{\text{\textbf{M}}}$, $\bar{\text{\textbf{R}}}$, and $\bar{\text{\textbf{C}}}$ indicate that only the fidelity dependence of the corresponding module is removed. Meanwhile, the symbol $\textbf{N}$ (None) indicates that all fidelity-dependent functions are not used, and training is performed only with a single high-fidelity dataset; the symbol \textbf{F} (full) indicates that all fidelity dependencies are introduced. In the experiments of this section, the training set is composed of a mixture of 100 randomly selected high-fidelity data frames and all low-fidelity data.

\begin{figure*}[htbp]
\centering
\includegraphics[width=\textwidth]{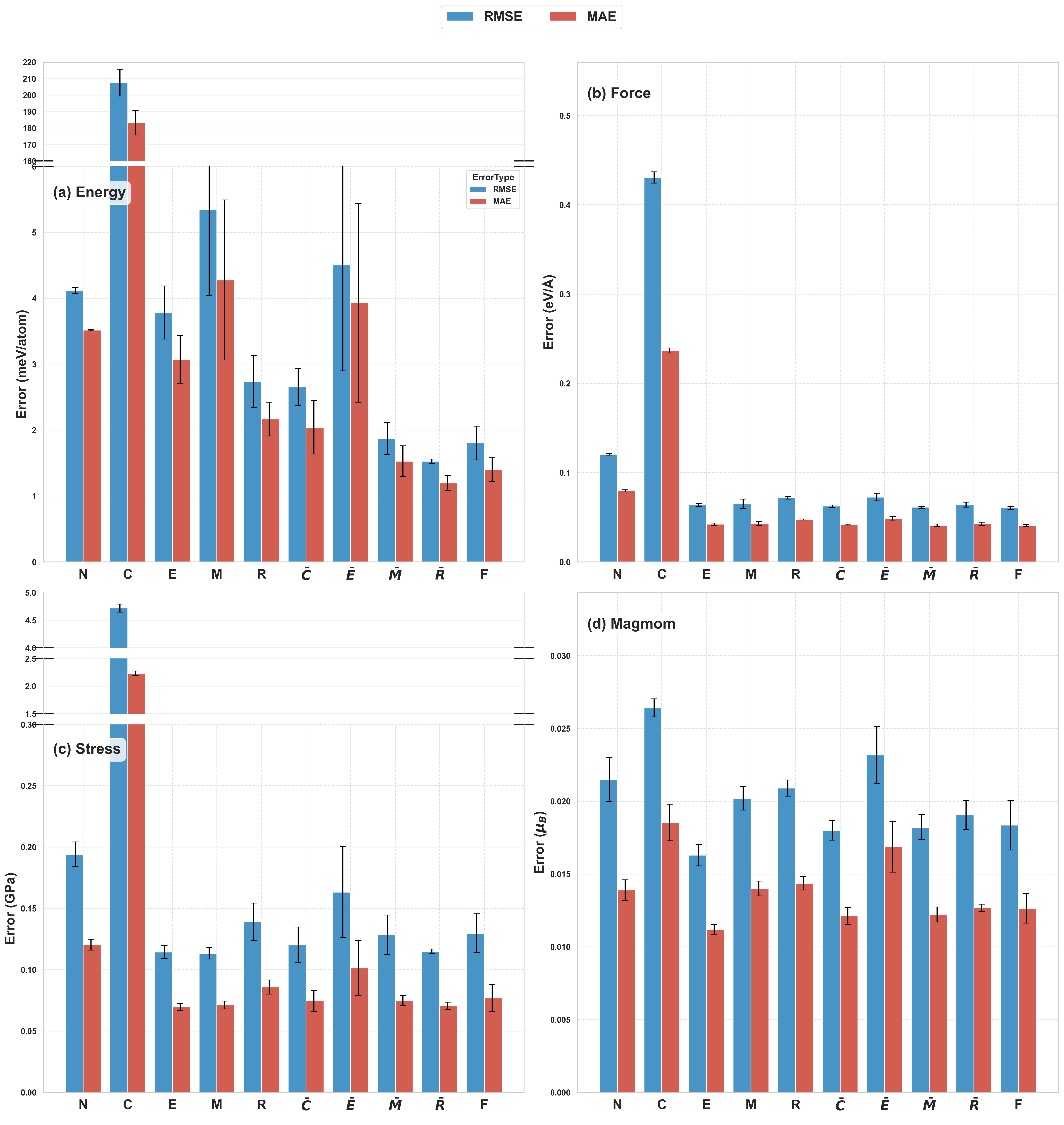}
\caption[experiment2]{Comparison of training accuracies with combinations of different fidelity-dependent functional modules. \textbf{N} denotes training using only a single high-fidelity dataset (100 frames); \textbf{F} denotes, on this basis, mixing all low-fidelity data and enabling all four modules for multi-fidelity force field training; \textbf{C}, \textbf{E}, \textbf{M}, and \textbf{R} denote respectively only enabling the corresponding fidelity-dependent composition model, atomic embedding layer, message-passing layer, and readout layer modules; $\bar{\text{\textbf{C}}}$, $\bar{\text{\textbf{E}}}$, $\bar{\text{\textbf{M}}}$, and $\bar{\text{\textbf{R}}}$ denote only disabling the corresponding functional modules.}
\label{function}
\end{figure*}

Figure \ref{function} compares the prediction accuracies of models obtained from different functional combinations. In terms of the comprehensive prediction accuracy for the four types of labels (total energy, atomic forces, lattice stress, and atomic magnetic moments), except for combination \textbf{C}, the results of all multi-fidelity-trained functional combinations outperform those of \textbf{N}. In particular, \textbf{F}, which enables all multi-fidelity functions, achieves the best results. This indicates that the combination of each functional module implemented in this work has the ability to perform multi-fidelity machine learning and improve the prediction accuracy of high-fidelity labels through low-fidelity data.

Notably, the generally poor results of \textbf{C} do not mean that the fidelity-dependent composition model is unimportant in this architecture. On the contrary, by comparing the total energy prediction results of $\bar{\text{\textbf{C}}}$ and \textbf{F}, it can be seen that when other functions are already included, optimizing the composition model can still further improve the prediction accuracy. This shows that the fidelity-dependent composition model is necessary but not sufficient for the multi-fidelity force field architecture. Comparing the total energy prediction results of \textbf{E}, $\bar{\text{\textbf{E}}}$, and \textbf{F} reveals that the fidelity-dependent atomic embedding module exhibits similar behavior.

Overall, the experiments in this section indicate that each fidelity-dependent module can exert a certain multi-fidelity machine learning function, and they need to be coordinated to achieve optimal performance.

\subsection{Comparison with Transfer Learning}

\begin{figure*}[htb]
\centering
\includegraphics[width=\textwidth]{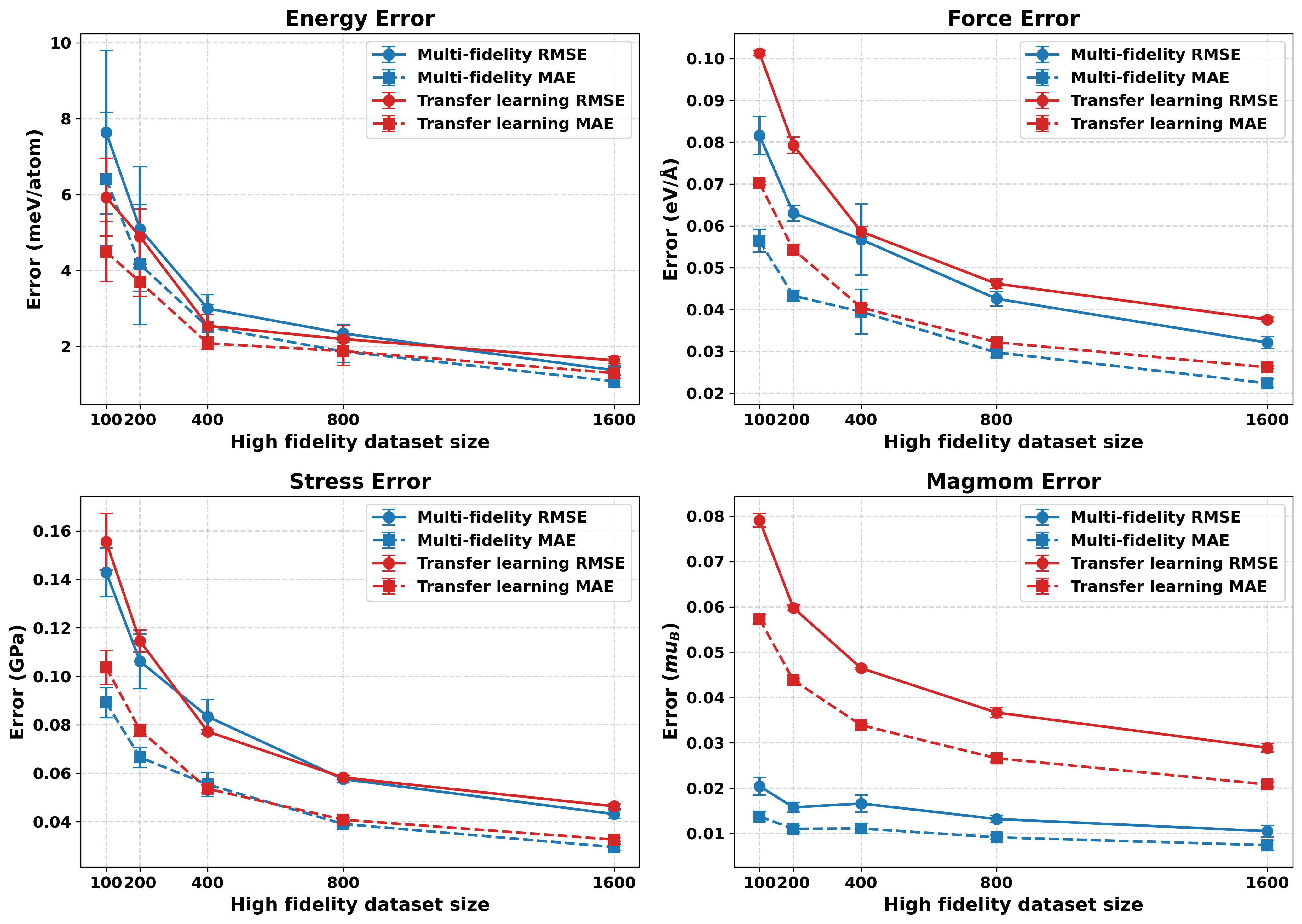}
\caption[experiment3]{Comparison of training effects between the multi-fidelity force field architecture and transfer learning. In this experiment, training is conducted using all low-fidelity data and a variable number of high-fidelity data (abscissa).}
\label{fig:transfer_learning_compare}
\end{figure*}

In this section, the multi-fidelity force field architecture implemented in this work is compared with previously commonly used multi-fidelity machine learning methods. Among them, the $\Delta$-learning method has relatively strict requirements for training data: it requires that a certain number of sample inputs have both high-fidelity and low-fidelity labels simultaneously, which limits its application in scenarios of force field training for cathode materials with scarce data. Therefore, this section mainly conducts comparative tests with transfer learning methods. Here, transfer learning is performed on the original CHGNet model. First, all low-fidelity datasets are used for weight pre-training, and the pre-trained model is saved. Second, a certain number of high-fidelity data are used to continue weight fine-tuning. In the fine-tuning stage, the weights of embedding layers and graph convolution layers related to atoms, bonds, and bond angles in the pre-trained model are frozen to preserve the general features of interatomic interactions learned by the model during pre-training. Finally, the reserved 10\% high-fidelity data are used for testing.

Figure \ref{fig:transfer_learning_compare} shows a comparison of test errors between the multi-fidelity force field architecture and transfer learning using the original framework. As can be seen from the figure, except that the error of total energy is similar, the performance of the multi-fidelity framework is superior to that of transfer learning training based on the original framework for most labels, including atomic forces, lattice stress, and atomic magnetic moments. Notably, in terms of performance on the task of predicting atomic magnetic moments, the multi-fidelity framework significantly outperforms transfer learning. Similar to the discussion in Section \ref{sec:effectiveness}, this indicates that the multi-fidelity architecture is capable of learning more general atomic representation features $\{\mathbf{v}_i^t\}$ from training data of different precisions.

\section{Conclusions}

This paper presents a multi-fidelity machine learning interatomic potential framework designed for cathode materials. By modifying multiple modules of the CHGNet architecture, effective integration of data with varying fidelity levels is achieved. The framework significantly alleviates the scarcity of high-quality training data while retaining the original magnetic moment prediction capability, thereby overcoming key obstacles in applying machine learning potentials to simulations of lithium-ion battery cathode materials. Experiments on LMFP materials demonstrate that incorporating low-fidelity non-magnetic data can effectively enhance the prediction accuracy of magnetic data labels, validating the model's ability to extract useful information from low-precision data. The framework supports a modular multi-fidelity learning mechanism, allowing flexible adaptation to different material systems, and outperforms conventional transfer learning and  $\Delta$-learning approaches.

The proposed method exhibits strong generalizability and can be extended to other graph neural network potentials. Future work may involve integrating more diverse and larger-scale cathode material datasets, extending to broader material systems, exploring more granular fidelity level divisions, and establishing quantifiable data quality evaluation standards. Additionally, a general multi-fidelity pre-trained model for battery materials could be developed based on public material databases. Such a model could be efficiently adapted to specific tasks or new material systems with minimal high-fidelity data fine-tuning, reducing reliance on extensive high-quality datasets and maximizing the value of open scientific data.




\end{document}